\documentclass[sigconf]{acmart}
\AtBeginDocument{%
  \providecommand\BibTeX{{%
    \normalfont B\kern-0.5em{\scshape i\kern-0.25em b}\kern-0.8em\TeX}}}

\setcopyright{acmcopyright}
\copyrightyear{2022}
\acmYear{2022}
\acmDOI{XXXXXXX.XXXXXXX}

\acmConference[FDG '22]{Foundations of Digital Games}{September 05--08,
  2022}{Athens, Greece}
\acmPrice{15.00}
\acmISBN{978-1-4503-XXXX-X/18/06}
\usepackage{multirow} 
\usepackage{subcaption}

\begin{document}


\title{Game~State Learning via Game~Scene~Augmentation}

\author{Chintan Trivedi}
\affiliation{%
  \institution{University of Malta, Msida, Malta}
  \streetaddress{University of Malta}
  \city{}
  \country{}}
\email{ctriv01@um.edu.mt}

\author{Konstantinos Makantasis}
\affiliation{%
  \institution{University of Malta, Msida, Malta}
  \streetaddress{University of Malta}
  \city{}
  \country{}}
\email{konstantinos.makantasis@um.edu.mt}

\author{Antonios Liapis}
\affiliation{%
  \institution{University of Malta, Msida, Malta}
  \streetaddress{University of Malta}
  \city{}
  \country{}}
\email{antonios.liapis@um.edu.mt}

\author{Georgios N. Yannakakis}
\affiliation{%
  \institution{University of Malta, Msida, Malta}
  \streetaddress{University of Malta}
  \city{}
  \country{}}
\email{georgios.yannakakis@um.edu.mt}
\renewcommand{\shortauthors}{Trivedi et al.}

\begin{abstract}
Having access to accurate game state information is of utmost importance for any artificial intelligence task including game-playing, testing, player modeling, and procedural content generation. Self-Supervised Learning (SSL) techniques have shown to be capable of inferring accurate game state information from the high-dimensional pixel input of game footage into compressed latent representations. Contrastive Learning is a popular SSL paradigm where the visual understanding of the game's images comes from contrasting dissimilar and similar game states defined by simple image augmentation methods. In this study, we introduce a new game scene augmentation technique---named \emph{GameCLR}---that takes advantage of the game-engine to define and synthesize specific, highly-controlled renderings of different game states, thereby, boosting contrastive learning performance. We test our \emph{GameCLR} technique on images of the CARLA driving simulator environment and compare it against the popular \emph{SimCLR} baseline SSL method. Our results suggest that \emph{GameCLR} can infer the game's state information from game footage more accurately compared to the baseline. Our proposed approach allows us to conduct game artificial intelligence research by directly utilizing screen pixels as input.
\end{abstract}



\maketitle

\section{Introduction}
\label{sec:introduction}
Extensive work \cite{yannakakis2018artificial,barthet2021go, berner2019dota} in the fields of dissimilar domains of AI and games such as player experience modeling, general gameplaying or content generation make use of the internal state of the game \cite{anand2019unsupervised, nelson2021estimates} obtained from the game engine. Using computer vision to obtain such state information from on-screen game footage, instead of directly from the game engine, remains challenging \cite{stooke2021decoupling}. Recent computer vision advancements with contrastive learning \cite{jaiswal2020survey}, however, show promise in tackling these challenges.

Contrastive learning belongs to the family of self-supervised representation learning methods in computer vision that use a ``pairwise-comparison'' approach which operates by contrasting semantically similar and dissimilar images. The pairwise mechanism helps the vision model to identify critical visual features that define the semantics of these images. Recent work \cite{trivedi2022representations} has applied this technique to the domain of learning state representations in games from pixel input. Such methods, however, rely on simple image augmentation techniques such as image flipping, rotation, and brightness change, to define and create semantically similar pairs of images. In this work, we investigate whether having access to a game engine can help us synthesize highly-controlled image augmentations that are better suited for learning such vision models. In particular, we use the game engine to construct better similar and dissimilar pairings via the proposed game scene augmentation technique, named \emph{GameCLR}, for the task of game state representation learning.

\begin{figure*}[!tb]
\includegraphics[width=\textwidth]{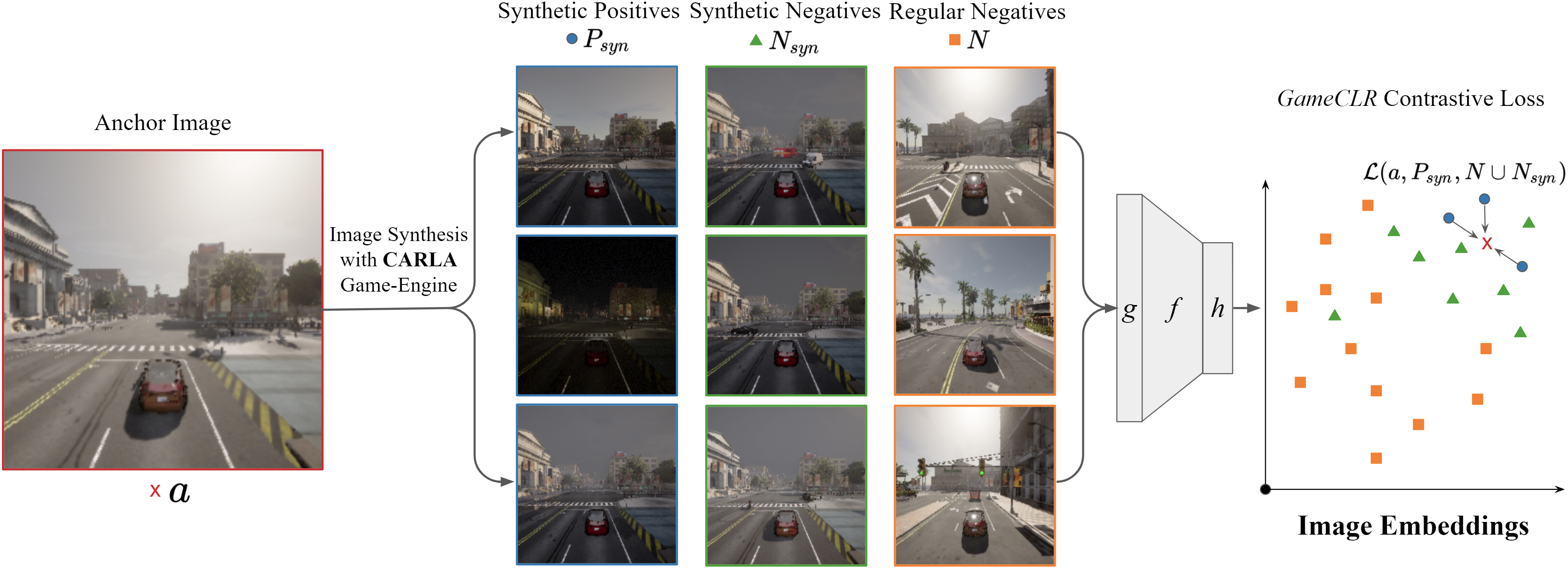}
\caption{The \emph{GameCLR} Contrastive Learning Framework.}
\label{fig:gameclr}
\end{figure*}

GameCLR synthesizes images that represent similar game states that are highly dissimilar in the pixel-space (\emph{synthetic positives}) and images that represent different game states but are very similar in the pixel-space (\emph{synthetic negatives}). Within the context of a car racing game, Fig. \ref{fig:gameclr} visualizes an example of such a representation space containing images that act as synthetic positives and negatives to a reference image called the \emph{anchor}. Our hypothesis is that by including such images in the contrastive learning process, our model will be better equipped to learn the important visual features that define any particular game state. Moreover, as we are defining the positive and negative pairings generated by the game engine, we also have a level of control over the learning process by guiding the model to learn what distinguishing factors are of importance to us (i.e. traffic information) and which ones can be considered as invariant for learning (e.g. rainy weather, color of the car). We test how training a vision model using GameCLR compares against a baseline SSL method SimCLR, on the CARLA driving simulator \cite{dosovitskiy2017carla}. Our findings suggest that synthesizing specific images from the game engine can boost the performance of contrastive learning methods for learning critical game state features from images.

\begin{table}[!tb]
\caption{Summary of the regular Image Augmentations and Game Scene Augmentations used in our paper.}
\centering
\begin{tabular}{|p{7cm}|l|}
\hline
\textbf{Image-based Augmentations:} & $g(X)$ \\
\quad \textit{Flipping, Noise, Change Brightness, Rotation, etc.} & \\
\hline
\textbf{Game Scene-preserving Augmentations:} & $e_p(S)$ \\
\quad Change weather (\textit{clear, cloudy, windy, wet, rainy}) & \\
\quad Change time of day (\textit{noon, sunset, midnight}) & \\
\quad Change ego-vehicle color (\textit{5 color options}) & \\
\hline
\textbf{Game Scene-altering Augmentations:} & $e_a(S)$ \\
\quad Add one, two, or three vehicles (one per lane) & \\
\hline
\end{tabular}
\label{tab:augmentations}
\end{table}

\section{Methodology}
\label{sec:dataset}

For all experiments reported in this study, we use the CARLA \cite{dosovitskiy2017carla} urban driving simulator (see Fig. \ref{fig:gameclr}) which provides access to its Unreal engine via a \emph{Python} API. A \emph{scene} $S$ in CARLA is defined as the current game state of its Unreal engine, which, when put through the game's graphic renderer $r$, yields the pixel output $X$ shown to the user on the screen (\emph{i.e.}, $X = r(S)$). We take advantage of this game engine to generate a dataset for testing two contrastive learning methods: (1) a baseline SSL method \emph{SimCLR} \cite{chen2020simple} which uses simple image augmentations $g(X)$; and (2) our proposed \emph{GameCLR} method which uses the CARLA game-engine to first apply game scene augmentation $S' = e(S)$ before going through rendering $X'=r(S')$ and then applying the regular image augmentations $g(X')$. All the augmentation techniques used across both methods are described in Table \ref{tab:augmentations}.

\subsection{SimCLR}
\label{subsec:simclr}
In 2020, Chen \emph{et al.} \cite{chen2020simple} proposed \emph{SimCLR}, a simple framework for contrastive learning of visual representations. A contrastive approach between similar and dissimilar images is used  to learn image representations based on the content present in the images. Its pipeline has four major components: an image augmentation function $g$ using simple image augmentations, a convolutional neural network encoder function $f$, a small fully-connected network called the projection head $h$ that maps representations to an embedding space, and a contrastive loss $\mathcal{L}$ that is applied on these embeddings. Under this framework, simple image augmentations (e.g. rotation, brightness, addition of noise, etc.) are used to create two different views of the same image that are semantically similar, referred to as a \emph{positive} pair. Similarly, any two views coming from distinct images are defined as \emph{negative} pairs due to semantic dissimilarity. For a given embedding $a$ of a reference image called the \emph{anchor}, and its positive pair's embedding $p$ as well as multiple negative pairs' embeddings in set $N$, the contrastive probability can be calculated as per Eq.~\eqref{eq:logit}.
\begin{equation}
\label{eq:logit}
        \mathcal{P}(a,p) = \frac{\text{exp}(a^T p/\tau)}{\text{exp}(a^T p/\tau) + \sum_{n \in N} \text{exp}(a^Tn/\tau)}
\end{equation}
\noindent where $\tau$ is the \emph{temperature} hyper-parameter and $a^T$ is the transposed $a$ matrix. Thus, the contrastive loss in SimCLR with respect to the anchor $a$ and all its associated positive pairs in a set $P$ can be defined as per Eq.~\eqref{eq:simclr_loss}:
\begin{equation}
\label{eq:simclr_loss}
    \mathcal{L}(a, P, N)=- \sum_{p \in P} \text{log} \mathcal{P}(a,p)
\end{equation}

We implement this SimCLR training method on our CARLA game dataset using the \emph{solo-learn} framework \cite{da2022solo}. We spawn the ego-vehicle at random locations and place the camera behind it. We also randomize the time of day, weather, color of the ego-vehicle, and traffic around it through the $e_a$ and $e_p$ functions described in Table \ref{tab:augmentations}. Through this process, we collect 50,000 anchor images to train a ResNet18 encoder \cite{da2022solo} over 20 epochs.

\begin{figure*}[ht]
     \centering
     \begin{subfigure}[b]{0.45\linewidth}
         \centering
         \includegraphics[width=\linewidth]{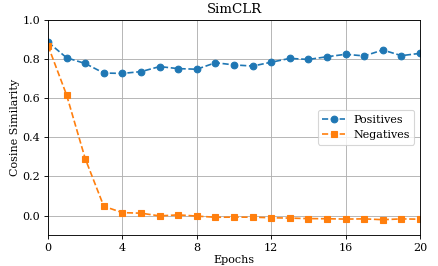}
     \end{subfigure}
     \begin{subfigure}[b]{0.45\linewidth}
         \centering
         \includegraphics[width=\linewidth]{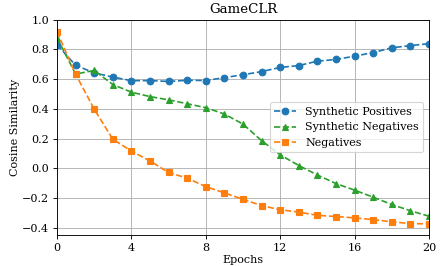}
     \end{subfigure}
        \caption{Average cosine similarity of the anchor image and its positive and negative pairings in a training batch.}
        \label{fig:cosine_sim}
\end{figure*}

\subsection{GameCLR (Our Approach)}
\label{subsec:gameclr}
Our work follows the literature \cite{kalantidis2020hard} regarding synthesizing \emph{hard negatives} which can provide more information to the SimCLR loss compared to regular negatives occurring through image augmentation. Our approach, however, exploits access to a game engine and thereby our ability to generate relevant images for learning meaningful representations. Our assumption in this paper is that we can accurately describe the traffic around the ego-vehicle without concern of changes in game aesthetics---such as car color---and lighting conditions arising from changes in weather and day time. 

Towards this end, we first render an anchor image by spawning the ego-vehicle at a random location. Then, we change the weather, time of the day conditions, or the color of the ego-vehicle while the ego-vehicle remains at the same state, using the Game Scene-preserving Augmentations $e_p(S)$ listed in Table \ref{tab:augmentations}. We define all such images as $P_{syn}$ indicating the set of synthetic positives with respect to the anchor image. Similarly, we synthesize negatives ($N_{syn}$) by spawning random vehicles around our ego-vehicle. This is done by performing Game Scene-altering augmentations $e_a(S)$ in addition to $e_p(S)$. Figure \ref{fig:gameclr} provides a few examples of the synthetic and regular images for a given anchor image. Note that all these images in GameCLR also undergo simple image augmentations during training, similar to SimCLR. Thus, we can now compute the GameCLR loss as $\mathcal{L}(a, P_{syn}, N \cup N_{syn})$ following the loss formulation of SimCLR in Eq.~\eqref{eq:simclr_loss}. This framework is showcased in Fig.~\ref{fig:gameclr} and we name it \emph{GameCLR}.
Our experiments for GameCLR follow similar choice of training hyper-parameters used in SimCLR as described in Section \ref{subsec:simclr}.

\section{Results}\label{sec:rl_results}

We present the results of our experiments with both augmentation approaches as a two-part assessment. First, we analyze in Section \ref{subsec:trainingeval} how the representations of game states change throughout the learning process, especially focusing on the behavior of the synthetic images used in GameCLR. Next, in Section \ref{subsec:posttrainingeval} we focus on highlighting the benefits of using such image representations for applications to game research that require extracting game state information from the game's images.

\subsection{Analyzing the Training Process}
\label{subsec:trainingeval}

To investigate the role of different images encountered in a training batch during the contrastive learning process in SimCLR, we start by measuring the average cosine similarities between the positive and negative pairs of image embeddings with respect to the anchor images in a given training batch. Figure \ref{fig:cosine_sim} (left) showcases those measurements during the training process across 20 epochs. At the beginning of training, both sets of positive and negative images have similar cosine similarity to the anchor images. This implies that the employed model before training cannot discriminate images that are semantically similar to an anchor image from images that are semantically different. As the training goes on, however, we notice that the images that belong to positive sets are more closely embedded to the anchor images compared to the images that belong to the negative sets. This behavior indicates that the model learns semantic similarities between images and embeds those semantics into the produced high-level image representations.

For GameCLR, to evaluate the degree to which game scene augmentation impacts the learning process, we measure the changes in cosine similarities between the synthetic positives, synthetic negatives and regular negatives with respect to the anchor throughout training (see Fig. \ref{fig:cosine_sim}). We observe that all sets of images start at a similar level of cosine similarity with the anchor, but as training advances, the synthetic negatives prove harder to contrast than regular negatives. Since both negatives are included in the contrastive loss of Eq. \eqref{eq:logit}, the higher and more granular loss provides a more informative learning signal to the model during training.
Interestingly, by the time the algorithm converges, the model learns to distinguish the synthetic negatives at a similar level as that of the regular negatives. This indicates that after convergence the model is easily able to distinguish between the distinct game states including the synthetic hard negatives. 

This analysis shows the superior learning capability afforded by the synthetic images obtained by directly modifying the pixels of the image with the help of the game engine. In order to quantify this benefit in terms of applicability to games research, we compare the models trained by these two approaches based on post-training evaluation, described in the following section.


\begin{table}[]
\caption{Average $R^2$ correlations between trained ResNet18 vectors and internal game state variables, averaged over 5 runs and shown along with 95\% confidence intervals. Highest average $R^2$ values for each variable are highlighted in bold.}
\begin{tabular}{|l|c|c|c|}
\hline
\textbf{Traffic variables} & \textbf{Untrained} & \textbf{SimCLR} & \textbf{GameCLR} \\ \hline
Dist. (left vehicle)&0.31$\pm$0.006&0.50$\pm$0.006&\textbf{0.60$\pm$0.010} \\
Dir. (left vehicle)&0.35$\pm$0.013&0.53$\pm$0.009&\textbf{0.56$\pm$0.008} \\ 
Dist. (front vehicle)&0.33$\pm$0.013&0.58$\pm$0.010&\textbf{0.70$\pm$0.014} \\
Dir. (front vehicle)&0.37$\pm$0.016&0.53$\pm$0.010&\textbf{0.57$\pm$0.007} \\
Dist. (right vehicle)&0.39$\pm$0.008&0.65$\pm$0.010&\textbf{0.69$\pm$0.005} \\
Dir. (right vehicle)&0.40$\pm$0.015&0.60$\pm$0.014&\textbf{0.65$\pm$0.009} \\ \hline
\end{tabular}
\label{tab:carla_correlations}
\end{table}

\subsection{Post-Training Evaluation}
\label{subsec:posttrainingeval}

As proposed by Anand \emph{et al.} \cite{anand2019unsupervised}, we evaluate how well the learned representations have captured information relevant to the game state through \emph{linear probing}. Linear probing includes freezing the weights of the ResNet encoder after the self-supervised training is over (i.e. the contrastive loss has converged). Then, we train linear regression models with the learned representations acting as the predictor variables (input) and certain variables describing the game state acting as the response variables (output). We measure the performance of these regression models with the $R^2$ correlation metric, where higher correlation values suggest that the model has better learned to identify the game state variables from the images.

In our study, we wish to test whether the derived representations of our models can describe the traffic around the ego-vehicle irrespective of weather and lighting conditions. Therefore, we prepare an evaluation dataset in CARLA by spawning an ego-vehicle at a random location with a camera and collecting RGB images, and at the same time collecting information about the coordinates and motion direction of the vehicles surrounding this ego-vehicle, similar to \cite{trivedi2022representations}. We refer to these as \textit{traffic variables}, as they can describe the state of traffic around the ego-vehicle. For each frame in our dataset, we collect a total of 6 synchronized traffic variables: \emph{Distance (left vehicle), Direction (left vehicle), Distance (front vehicle), Direction (front vehicle), Distance (right vehicle), Direction (right vehicle)}. Note that we are able to find this ground truth of the traffic variables due to direct access to the game engine of CARLA. Let us stress that the traffic variables are not used during the training of our contrastive models; they are only used as desired output for linear probing after training is completed.

In Table \ref{tab:carla_correlations}, we present the average correlation values observed for each of the 6 game state variables present in our evaluation dataset.  First, we observe that both methods---SimCLR and GameCLR---improve upon the baseline of a randomly initialized ResNet18 model, verifying that contrastive learning is an effective solution for learning to differentiate between distinct game states. Across the six in-game variables, SimCLR provides a 157\% improvement over the untrained baseline, on average, whereas GameCLR provides an improvement of 174\% on average. 

Since contrastive learning is guided by engine-specific hard negatives in GameCLR, the representations obtained by this method outperform SimCLR by approximately 11\% on average on the linear probing task while using the same amount of images and training steps. This suggests that the ResNet18 encoder trained using the GameCLR approach extracts more meaningful representations that better capture traffic information in the game image as compared to SimCLR. All $R^2$ values for the different in-game variables in GameCLR are significantly higher than SimCLR ($p<0.05$), with the highest improvement achieved for the distance to left vehicle (20\% improvement over SimCLR), and---surprisingly---the least improved $R^2$ was for the direction to the left vehicle (5.5\% improvement over SimCLR).

\section{Conclusion}
In this paper we introduced GameCLR, a contrastive learning technique for learning game state representations. The main contribution of this technique is the introduction of game engines for synthesizing training images and enriching data augmentation in this fashion. We notice that by synthesizing hard positives and negatives for each associated anchor image, we can better guide the contrastive learning process. Our results in the driving simulator CARLA suggest a 11\% average improvement (in terms of $R^2$) when extracting critical traffic-related game state features from images of this game with our GameCLR approach over another comparable approach (SimCLR). Our proposed method enables the user to control which visual features of a game the SSL method learns from the input RGB images by specifying which engine variables (that impact rendering) produce synthetic positives and which produce synthetic negatives. Moreover, it shows the performance improvement over the standard contrastive learning approach SimCLR which uses simple image-based augmentation methods and does not exploit the game engine, as traditionally done when such computer vision methods are applied to games. Our proposed method enables the use of the game's images as input instead of explicit state information, with downstream applications in AI and games research like deep reinforcement learning for game-playing, pixel-level procedural content generation or correlating affect with game-play footage in player modeling.

\section*{Acknowledgement}

This work was supported by the European Union’s H2020 research and innovation programme [Grant Nos. 951911, 101003397].

\bibliographystyle{ACM-Reference-Format}
\bibliography{contrastive_game_engine.bib}
\end{document}